\documentclass[runningheads]{llncs}

\usepackage{eccv}

\usepackage{eccvabbrv}          
\usepackage{graphicx}
\usepackage{booktabs}
\usepackage{multirow}
\usepackage{amsmath}
\usepackage{amssymb}
\usepackage{pifont}
\usepackage{algorithm}
\usepackage{algpseudocode}             
\floatstyle{ruled}\restylefloat{algorithm}
\algrenewcommand\algorithmicrequire{\textbf{Input:}}
\algrenewcommand\algorithmicensure{\textbf{Output:}}
\algrenewcommand{\algorithmiccomment}[1]{\hfill$\triangleright$~\textit{#1}}

\usepackage{hyperref}
\usepackage{orcidlink}
\usepackage{placeins}

\newcommand{\hota}{\textsc{HOTA}}
\newcommand{\deta}{\textsc{DetA}}
\newcommand{\assa}{\textsc{AssA}}
\newcommand{\loca}{\textsc{LocA}}

\newcommand{\cmark}{\ding{51}}
\newcommand{\xmark}{\ding{55}}

\begin{document}
\title{Online Multi-Camera 3D Tracking via\texorpdfstring{\\}{ }%
ID Prediction over Recurrent Sparse Queries}
\titlerunning{ID Prediction over Recurrent Sparse Queries}

\author{Pragyan Shrestha \and
        Haruto Nakayama \and
        Atom Scott}
\authorrunning{P.~Shrestha \etal}
\institute{Playbox Inc., Tokyo, Japan\\
           \email{\{pragyan,haruto.nakayama,atom\}@playbox.co}}

\maketitle

\begin{abstract}
Online multi-camera 3D tracking must maintain scene-global identities across synchronized views, yet query-based trackers carry these identities only implicitly in the instance bank, where they fragment upon query interruption. We present an online architecture that recovers association accuracy by predicting IDs explicitly over recurrent sparse queries. An outside-in Sparse4D detector fuses calibrated views into world-frame 3D detections while propagating a sparse query bank, and a causal MOTIP ID decoder associates detections against a finite trajectory memory. We adapt MOTIP's relative-ID prediction and recycled-slot runtime to globally fused 3D observations, and introduce metric spatial gating and proximity-based newborn recovery. On the official 2026 AI City Challenge Track 1 test set, our method raises \hota{} from 29.63 with native instance-bank identities to 38.01, primarily through an \assa{} increase from 20.83 to 31.10, and ranks third on the public leaderboard. Full-sequence validation over all 9{,}000 frames of each scene shows that decoupled ID training improves \hota{} over native identities, whereas continuing detector training alongside the detached ID objective produces scene-dependent gains and losses.
\end{abstract}

\section{Introduction}
\label{sec:intro}

Large indoor facilities increasingly rely on networks of calibrated cameras
for live monitoring. Their central perception problem is 3D Multi-Target
Multi-Camera (MTMC) tracking. At every frame, the system must recover each
agent's 3D bounding box and maintain one identity across cameras and time.
Track~1 of the 2026 AI City Challenge instantiates this setting as
seven-class 3D tracking from synchronized, calibrated cameras in warehouse
scenes \cite{Tang26AICity26}. Unlike commonly used pedestrian-only benchmarks
such as WILDTRACK and MultiviewX \cite{wildtrack,mvdet}, the challenge
requires metric 3D boxes and scene-global identities for multiple object
classes.

A common MTMC design first detects and tracks objects independently in each
camera and then associates the resulting tracklets across views using
appearance and spatio-temporal constraints
\cite{lmgp}. This late-aggregation decomposition
is modular, but fragmented tracklets and identity errors produced by the
per-camera trackers must be resolved during cross-camera fusion.
Other methods reduce this separation through global spatial-temporal
association \cite{rest,gmt}.
Early-aggregation methods instead combine calibrated views in a shared
bird's-eye-view or 3D representation before temporal association
\cite{earlybird,tracktacular,depthtrack,mcblt}.

Query-based camera-only 3D trackers provide another alternative by carrying
object-centric state between frames
\cite{mutr3d,pftrack,dqtrack,adatrack}. Several of these methods add learned
association or extra query persistence under occlusion, so a missed
detection does not necessarily destroy an identity. Sparse4D v3 instead
attaches native IDs to cached recurrent queries in a bounded instance bank
\cite{sparse4dv3}. After a query leaves the bank, that assignment provides
no explicit mechanism for recovering its previous scene-global identity.
MOTIP \cite{motip} addresses the analogous association problem in
single-camera 2D tracking by predicting each current detection's identity
from an in-context window of historical trajectory tokens.

\begin{figure}[t]
  \centering
  \includegraphics[width=\linewidth]{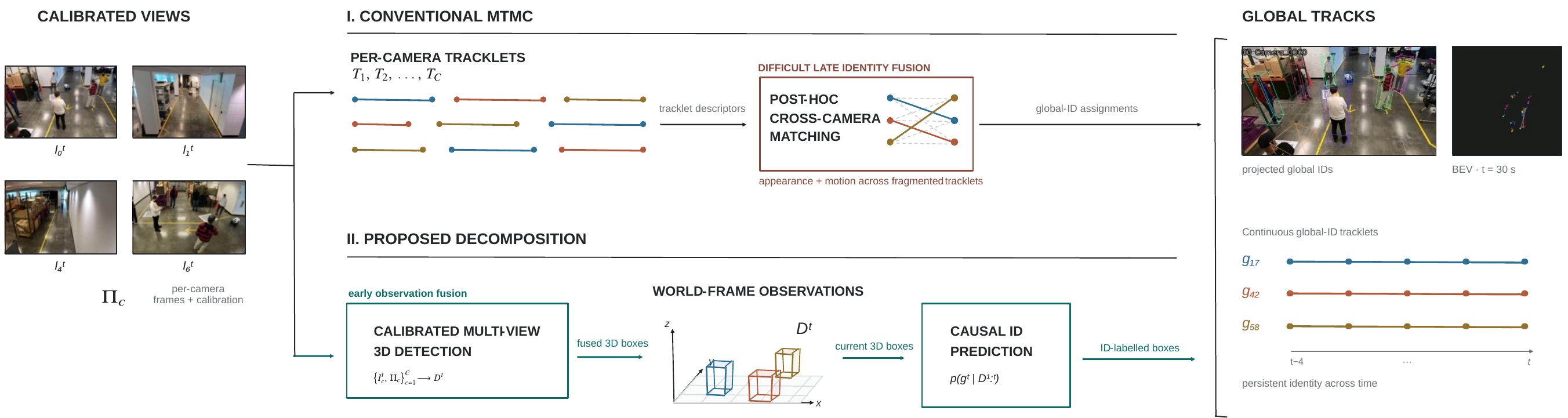}
  \caption{\textbf{Difference in multi-camera tracking fusion strategies.} Conventional MTMC first
  forms per-camera tracklets and then performs difficult post-hoc
  cross-camera fusion. Our decomposition instead fuses calibrated views into
  one world-frame 3D observation stream before online ID prediction.}
  \label{fig:overview}
\end{figure}

We transfer this formulation to online multi-camera 3D tracking
(Fig.~\ref{fig:overview}). Our detector follows the outside-in Sparse4D
adaptation of Wang et al.\ \cite{nvidia_omni3d}, building on recurrent sparse
temporal fusion \cite{sparse4dv2,sparse4dv3}. It fuses all cameras into
world-frame 3D detections and propagates sparse instance queries between
frames. We retain this recurrent detection path and compare against its
native instance-bank ID assignment, while a separate ID-prediction decoder
maintains the longer-lived association state. The decoder associates each
accepted detection against a sliding memory of past trajectory tokens.
Following MOTIP, a bounded relative-ID vocabulary supports in-context
classification, while recycled labels are mapped to monotonically allocated
output IDs. We adapt this runtime to scene-global 3D tracks. Identity can
therefore survive a short interruption even after its detector query leaves
the instance bank.

World-frame geometry also constrains assignment. We mask identity claims
that require physically implausible motion and reconnect newborn predictions
that occur beside a recently silent track.
These geometric constraints are independent of the learned token representation.
We evaluate explicit 3D position in the token, but do not assume that it is
complementary to the detector query feature.

The online system placed third on the 2026 full-test public leaderboard. On
the official test server, ID prediction with geometric constraints raises
the recurrent detector's native tracking from
29.63 to 38.01~\hota{}, improving association while detection remains
essentially unchanged. Validation ablations show that continuing detector
training alongside the detached ID objective is scene-dependent. It helps the
scene the pretrained detector handles worst and harms the one it handles best.
We further ablate the ID-token input and the geometric constraints.

Our central contribution is the adaptation and evaluation of causal ID
prediction over recurrent, globally fused multi-camera 3D detections.
Unlike post-hoc cross-camera trajectory fusion, association acts
directly on a single world-frame observation stream. Unlike native query
propagation, the identity layer has its own trajectory memory. Our additions
to the MOTIP runtime are metric spatial gating and proximity-based newborn
recovery. We evaluate the design over each full 9{,}000-frame validation
sequence, comparing against native instance-bank identities and ablating
detector training, token composition, and geometric constraints.

\section{Related Work}
\label{sec:related}

\noindent\textbf{Multi-target multi-camera tracking.}
A common late-aggregation MTMC pipeline performs detection and tracking
separately in each camera, then associates tracklets across cameras using
appearance, camera geometry, and temporal consistency
\cite{lmgp}. Such pipelines are vulnerable to
fragmentation and identity errors produced before cross-camera fusion. For
online association, ReST \cite{rest} matches detections spatially across
cameras before temporal graph association, while GMT \cite{gmt} directly
matches new detections to multi-view global trajectories. For
overlapping cameras, MVDet \cite{mvdet} and MVDeTr \cite{mvdetr} instead
aggregate multi-view evidence on a bird's-eye-view plane for pedestrian
detection. EarlyBird \cite{earlybird}, TrackTacular
\cite{tracktacular}, and DepthTrack \cite{depthtrack} extend BEV aggregation
to temporal tracking. Most closely
related to our decomposition, MCBLT \cite{mcblt} first produces globally
fused 3D BEV detections and then performs long-sequence association with
hierarchical graph neural networks. We likewise associate a single globally
fused 3D observation stream, but use online in-context ID prediction rather
than hierarchical graph optimization.

\smallskip\noindent\textbf{Recurrent query-based 3D tracking.}
DETR-style detectors \cite{detr,deformabledetr} represent objects as sparse
queries. In 2D tracking, MOTR \cite{motr} and TrackFormer
\cite{trackformer} propagate identity-preserving track queries, whereas
MeMOT \cite{memot} stores track embeddings in an explicit spatio-temporal
memory. Camera-based 3D trackers subsequently extended query propagation
across multiple views. MUTR3D \cite{mutr3d} introduces multi-view 3D track
queries, PF-Track \cite{pftrack} uses historical queries and motion prediction
for occlusion handling, and DQTrack \cite{dqtrack} separates object and track
queries. ADA-Track \cite{adatrack} alternates multi-view detection with
learned query association using appearance and geometry. Sparse4D
\cite{sparse4dv1,sparse4dv2,sparse4dv3} instead performs sparse deformable
multi-view aggregation and recurrently carries instance features between
frames. Its outside-in adaptation \cite{nvidia_omni3d} places these queries in
a shared world frame for fixed camera networks and supplies the detector used
in our system.

\smallskip\noindent\textbf{ID prediction and geometric constraints.}
MOTIP \cite{motip} decouples object detection from association and formulates
the latter as in-context ID prediction. Its decoder classifies current
detections using recent trajectory tokens labeled with relative identities.
Its runtime also recycles concluded relative labels and maps them to
monotonically allocated output IDs. We retain these mechanisms while
transferring MOTIP from single-camera 2D tracking to globally fused
multi-camera 3D detections. Geometric rejection is well established in
tracking-by-detection. DeepSORT \cite{deepsort} uses Mahalanobis gating,
ByteTrack \cite{bytetrack} associates every detection including low-score
boxes, and OC-SORT
\cite{ocsort} improves motion estimation under occlusion. Track-aware
initialization has also been used to suppress spurious newborn tracks
\cite{tracktrack}. Our spatial gate and proximity-based newborn recovery
adapt these ideas to metric world-frame distances after learned ID prediction.

\section{Method}
\label{sec:method}

\subsection{Problem Formulation and Tracker State}

At frame $t$, $C$ synchronized cameras provide images
$\mathcal{I}^t=\{I_c^t\}_{c=1}^{C}$ and fixed calibrations
$\{\Pi_c\}_{c=1}^{C}$. The required output is a set of world-frame oriented
boxes $b_i^t=(x,y,z,w,l,h,\theta)$, class labels $y_i^t$, confidence scores,
and scene-global identities $g_i^t$. The tracker is strictly causal. Output
at $t$ may depend on frames $1{:}t$, but never on a future frame.

\begin{figure}[t]
  \centering
  \includegraphics[width=\linewidth]{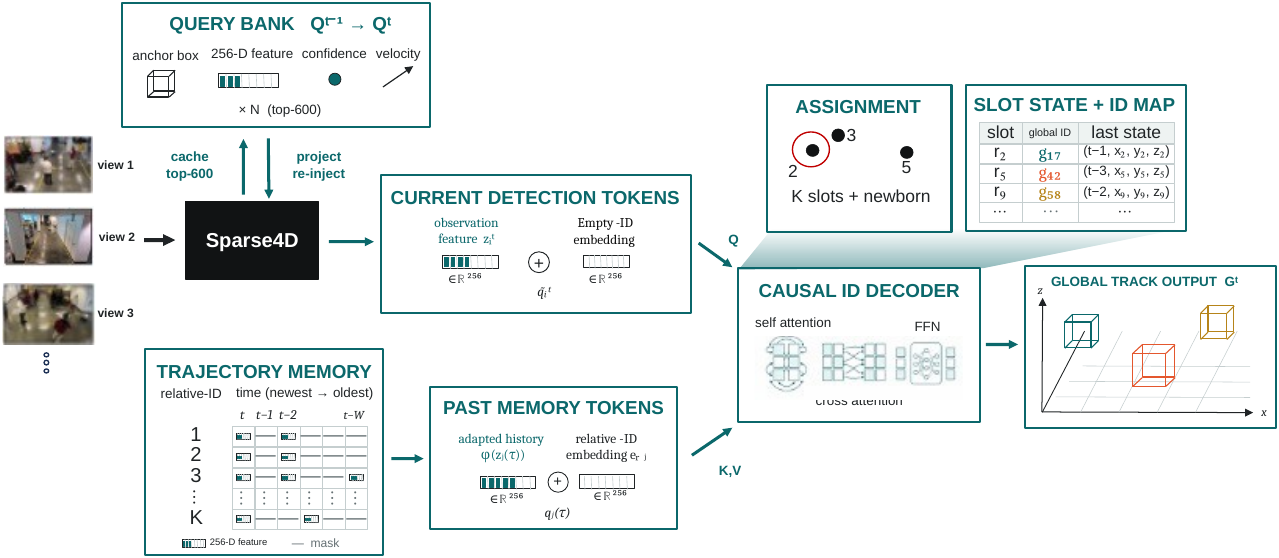}
  \caption{\textbf{Architecture details.} The recurrent detector
  fuses all calibrated views with query bank $\mathcal{Q}^{t-1}$ to produce
  world-frame observations and $\mathcal{Q}^{t}$. Detections query the
  finite trajectory memory $\mathcal{M}^{t-1}$ for relative
  identities.
  Geometric masking, discrete assignment, newborn handling, and the
  MOTIP least-recently-used (LRU) slot lifecycle map those labels through
  $\mu^t$ to scene-global IDs. Only
  $\mathcal{Q}^{t}$, $\mathcal{M}^{t}$, and $\mu^t$ cross the frame
  boundary.}
  \label{fig:method_architecture}
\end{figure}

Our state separates detection continuity from identity continuity.
\begin{equation}
  \mathcal{S}^t =
  \bigl(\mathcal{Q}^t,\mathcal{M}^t,\mu^t\bigr).
  \label{eq:state}
\end{equation}
Here $\mathcal{Q}^t$ is Sparse4D's recurrent instance bank,
$\mathcal{M}^t$ is a finite window of accepted trajectory tokens, and
$\mu^t$ maps the decoder's bounded relative-ID slots to monotonically
allocated global IDs. The number accumulated over a scene is unbounded,
although at most $K$ slots can be active simultaneously. A frame update
factors as
\begin{align}
  (\mathcal{D}^t,\mathcal{Q}^t)
    &= F_\theta(\mathcal{I}^t,\{\Pi_c\},\mathcal{Q}^{t-1}), \\
  (\mathcal{G}^t,\mathcal{M}^t,\mu^t)
    &= A_\psi(\mathcal{D}^t,\mathcal{M}^{t-1},\mu^{t-1}),
  \label{eq:update}
\end{align}
Here $F_\theta$ is the recurrent multi-view detector and $A_\psi$ the
ID-prediction runtime. Fig.~\ref{fig:method_architecture} details their
recurrent state updates.

\subsection{Recurrent Multi-View 3D Detection}
\label{sec:method_detector}

We adopt the outside-in Sparse4D detector
\cite{nvidia_omni3d,sparse4dv3}. A shared ResNet-101 and FPN encode every
camera. Nine hundred sparse 3D anchor queries sample the projected
multi-camera feature maps and are repeatedly refined by deformable
aggregation, self-attention, and box/class heads. Because fusion occurs
before tracking, overlapping camera observations produce one world-frame
detection stream rather than separate per-camera tracklets.

For every accepted detection, the final decoder layer emits
\begin{equation}
  d_i^t=(b_i^t,y_i^t,p_i^t,f_i^t),
\end{equation}
where $p_i^t$ is the detection confidence and
$f_i^t\in\mathbb{R}^{256}$ is its sparse query feature. The instance bank
retains the top 600 queries and re-injects them at the next timestamp after
projecting their anchor centers with predicted velocity. We reset the bank
only at a scene boundary and otherwise carry it through the complete
sequence.

The instance bank is deliberately not the final identity store. Its query
IDs are allocated monotonically and remain attached to cached queries, but a
query can be removed when its confidence falls. A later reappearance then
receives a new native ID. The association layer therefore consumes detector
boxes and features as observations and maintains identity state separately.

\subsection{From Sparse Detections to Trajectory Tokens}
\label{sec:method_tokens}

For each detection, we construct one of three observation features.
\begin{equation}
  z_i^t \in
  \bigl\{f_i^t,\ \gamma(c_i^t),\ f_i^t+\gamma(c_i^t)\bigr\},
  \label{eq:feature}
\end{equation}
corresponding to query-only, position-only, and combined tokens. The submitted
configuration uses the combined form with a 256-dimensional Fourier encoding
$\gamma$. It applies sine and cosine at 42 geometrically spaced frequencies
from 1 to 10 to each metric $(x,y,z)$ coordinate and zero-pads the resulting
252 values. Section~\ref{sec:exp_components} compares the three forms. We call
$f_i^t$ a \emph{detector query feature}, rather than a pure appearance
embedding. It is optimized for classification and box regression and can
already contain spatial information.

Following MOTIP, a residual feed-forward trajectory adapter $\phi$ transforms
stored history features, while current features bypass it. Each stored
observation is paired with the learned embedding of a relative identity slot
$r\in\{0,\ldots,K-1\}$.
\begin{equation}
  q_i^\tau = [\,\phi(z_i^\tau) \mathbin{\|} e_r\,]\in\mathbb{R}^{512}.
  \label{eq:token}
\end{equation}
Each current detection uses the same observation feature but enters the
decoder with
$\tilde q_i^t = [\,z_i^t \mathbin{\|} e_{\varnothing}\,]$,
where $e_{\varnothing}$ is a dedicated empty-ID embedding. Consequently, the
decoder must infer identity from the trajectory context rather than read it
from the current token.

\subsection{Causal ID Prediction in 3D}
\label{sec:method_motip}

Following MOTIP \cite{motip}, association is formulated as in-context
classification. The memory $\mathcal{M}^{t-1}$ stores accepted tokens from the
preceding $W=29$ frames, one fewer than the 30-frame training clip so that the
learned relative-time bias is never indexed beyond its range. A binary mask
distinguishes a missing observation from an inactive trajectory column.

Six decoder layers cross-attend from the current detections to the flattened
trajectory memory. A causal mask blocks tokens at the current or a future
timestamp, and each attention head receives a learned bias indexed by token
age. From the second layer onward, self-attention among the current
detections lets them resolve competing claims. The final head produces
\begin{equation}
  \ell_i^t\in\mathbb{R}^{K+1}, \qquad K=128,
\end{equation}
covering $K$ relative-ID slots and a separate \emph{newborn} label. The
relative labels make the classification problem bounded even though the
number of global identities accumulated over a scene is not.

\subsection{Geometry-Constrained Assignment and ID Lifecycle}
\label{sec:method_runtime}

\noindent\textbf{Spatial feasibility.}
Let $u_i^t=(x_i^t,y_i^t)$ be a detection's ground-plane center, and let
$(t_r,\hat u_r)$ be the last observation of relative slot $r$. Before
discrete assignment, the probability of slot $r$ is masked when
\begin{equation}
  \lVert u_i^t-\hat u_r\rVert_2 >
  \rho_{\mathrm{gate}}+\eta\max(t-t_r,1),
  \label{eq:gate}
\end{equation}
with $\rho_{\mathrm{gate}}=2.5$\,m and
$\eta=0.15$\,m/frame. This is an output constraint on the ID classifier. It
does not require position to be present in the learned token.

\smallskip\noindent\textbf{Newborn recovery.}
A briefly missed object may reappear with a degraded feature and be labeled
newborn, creating a duplicate identity beside a silent trajectory. For every
unclaimed newborn, we collect every unclaimed live identity inside
$\rho_{\mathrm{new}}+\eta\max(t-t_r,1)$, with
$\rho_{\mathrm{new}}=0.8$\,m, and assign greedily by increasing distance.
Each detection and each silent identity is used at most once. A newborn
with no such neighbor opens a new identity.

\smallskip\noindent\textbf{Assignment and lifecycle.}
Algorithm~\ref{alg:runtime} summarizes the complete causal update for one
frame, with $\theta_{\mathrm{id}}=0.2$ and $\theta_{\mathrm{new}}=0.6$.
An identity remains live while at least one unmasked token survives
in the window. Following MOTIP, relative slots are reusable implementation
labels, and only the global IDs produced through $\mu^t$ are written to the
challenge output. An LRU queue moves assigned slots to the back, while
newborns receive the oldest currently unclaimed slots. Reuse purges the old
tokens and remaps $\mu^t$ to fresh global IDs. If existing claims and newborns
exceed $K$ in one frame, excess newborn detections are discarded. This
preserves unique labels but bounds concurrent accepted identities by $K=128$.
All tracker state is reset once per scene.

\begin{algorithm}[t]
\caption{Causal ID assignment and lifecycle (one frame $t$)}
\label{alg:runtime}
\small
\begin{algorithmic}[1]
\Require detections $\mathcal{D}^t$, memory $\mathcal{M}^{t-1}$,
         slot map $\mu^{t-1}$, LRU queue
\Ensure  global IDs $\mathcal{G}^t$, memory $\mathcal{M}^{t}$,
         slot map $\mu^{t}$
\Statex
\State $\ell_i \gets$ ID-decoder logits over $K$ slots $+$ newborn
       \Comment{Sec.~\ref{sec:method_motip}}
\State mask every pair $(i,r)$ violating the spatial gate
       \Comment{Eq.~\eqref{eq:gate}}
\State object-max assignment, each detection and slot used at most once
\Statex
\State demote claims with slot probability $<\theta_{\mathrm{id}}$ to newborn
\State discard newborns with detector score $p_i^t<\theta_{\mathrm{new}}$
\State resume the nearest unclaimed live slot within
       $\rho_{\mathrm{new}}+\eta\max(t-t_r,1)$
\Statex
\State discard excess newborns if fewer than $K$ slots remain this frame
\State assign LRU slots to remaining newborns, then purge and remap $\mu^{t}$
\State $\mathcal{G}^t_i \gets \mu^{t}(r)$ for every accepted assignment $(i,r)$
\State append accepted tokens to $\mathcal{M}^{t}$, trim to window $W$,
       and drop empty slots
\end{algorithmic}
\end{algorithm}

\subsection{Training Strategy}
\label{sec:method_training}

For ID-decoder training, we process 30-frame clips at random temporal stride
1--4 through the recurrent detector. Detector outputs are Hungarian-matched to
ground-truth boxes and classes, and only matched query features construct the
ID-training tokens.
For each clip, six augmentation groups independently map tracks to random
relative IDs. First observations are newborn. With probability 0.5 per track,
temporal occlusion masks a random contiguous span, and with probability 0.5
per slot, identity-switch augmentation exchanges trajectory observations
within a frame. These augmentations prevent copying a fixed slot layout, and
every decoder layer receives cross-entropy supervision.

Our primary model freezes the detector and optimizes only the ID-prediction
modules. This isolates association learning from changes in detection. For
the detector-adaptation analysis, we first converge the ID components and
then continue training the recurrent detector and ID modules at lower learning
rates. The objectives remain decoupled. ID-loss gradients are detached from
the detector.

\section{Experiments}
\label{sec:exp}

\subsection{Setup}
\label{sec:exp_setup}

The 2026 AI City Challenge Track~1 dataset \cite{Tang26AICity26} contains 20 training, three
validation, and five test warehouse scenes, with synchronized 1080p video at
30\,fps, calibrated cameras, and seven agent classes. We add 15 warehouse
scenes from the 2025 edition \cite{aicity2025} for training after
checking that their camera rigs do not overlap the 2026 validation rigs.
For ablation studies, we use the validation scenes from the 2026 edition. \texttt{Warehouse\_020} and
\texttt{Warehouse\_021} share identical calibration and ground truth but use different appearance renderings.
\texttt{Warehouse\_022} is a distinct four-camera scene. We report them
separately rather than averaging the duplicated scenario.

The challenge evaluates 3D boxes and scene-global identities with per-class
3D HOTA aggregated across scenes \cite{Tang26AICity26}. Each validation sequence
contains 9{,}000 frames. We report the HOTA family (\hota{}, \deta{}, \assa{},
and \loca{}) \cite{hota} over every complete sequence using the official
NVIDIA 3D-IoU evaluator.

The detector starts from NVIDIA's synthetic-warehouse Sparse4D checkpoint
\cite{nvidia_omni3d}. Its optional visibility-weighted ReID branch is disabled
in the submitted configuration. The ID decoder therefore receives the final
256-dimensional detector query feature described in
Sec.~\ref{sec:method_tokens}, rather than a ReID-trained appearance embedding.
Following the official MOTIP codebase, we retain its trajectory-modeling
module. Immediately before ID decoding, it applies residual feed-forward
transformations to the stored history, while current detection features
bypass it.
All images are resized to $540\times960$. The ID decoder is optimized for
6{,}000 steps with AdamW, learning rate $4{\times}10^{-4}$, weight decay
$10^{-3}$, 200 warm-up steps, and gradient clipping at 1.0. Training uses bf16
on eight A100 GPUs with one recurrent clip per GPU. The continued-detector
variant resumes the converged ID model for 5{,}000 steps with detector/ID
learning rates of $2{\times}10^{-5}/10^{-4}$. Both modules update, but the
ID-loss gradient remains detached from the detector.

\smallskip\noindent\textbf{Runtime.}
On one NVIDIA RTX PRO 6000 Blackwell Max-Q GPU and a Threadripper 9970X,
the TensorRT detector with the PyTorch ID layer processes the 16-camera scenes
at 30.8--32.1\,FPS and the four-camera scene at 58.6\,FPS. These
9{,}000-frame averages exclude JPEG decoding and host-to-device transfer.
Against all-PyTorch inference at 6.7 and 22.9\,FPS, respectively, this is a
2.6--4.8$\times$ speedup. The ID layer adds 5.5--7.3\,ms per frame, while
TensorRT reduces validation \hota{} by 0.40--1.78 points across the scenes.

\smallskip\noindent\textbf{Test submission.} The leaderboard entry
(Table~\ref{tab:leaderboard}) was produced by the frozen-detector system
(NGC-initialized frozen detector $+$ MOTIP) run online over the five test scenes at
detection threshold 0.4 with the geometric constraints of
Sec.~\ref{sec:method_runtime}.

\subsection{Challenge Result}
\label{sec:exp_main}

\begin{table}[t]
  \centering
  \caption{\textbf{Public leaderboard of Track~1, 2026 AI City Challenge}
  (full test set, snapshot from August 15, 2026, top five of ten shown).
  The last row is our baseline,
  the recurrent Sparse4D tracker with its native instance-bank identities
  \cite{nvidia_omni3d}, evaluated through the same official server.}
  \label{tab:leaderboard}
  \begin{tabular}{@{}clcccc@{}}
    \toprule
    Rank & Team & \hota{} & \deta{} & \assa{} & \loca{} \\
    \midrule
    1 & EVA                    & 56.54 & 55.64 & 49.39 & 79.56 \\
    2 & SKKU-AL-T1             & 52.01 & 45.31 & 56.50 & 76.24 \\
    3 & \textbf{Playbox (ours)}& \textbf{38.01} & 40.26 & 31.10 & 75.18 \\
    4 & VTX\_QDTers            & 34.18 & 29.47 & 33.11 & 17.68 \\
    5 & TU-YMLab               & 25.88 & 26.52 & 30.62 & 69.70 \\
    \midrule
    -- & Sparse4D \cite{nvidia_omni3d} & 29.63 & 39.57 & 20.83 & 75.17 \\
    \bottomrule
  \end{tabular}
\end{table}

Table~\ref{tab:leaderboard} reports the public test result. The baseline row
uses the same recurrent detector but reports its native, monotonically
allocated instance IDs, which remain attached to queries only while cached.
Both entries were scored by the official evaluation server on the full test
set. Adding ID prediction with the geometric constraints
raises \hota{} from 29.63 to 38.01, with the gain concentrated in
association (\assa{} $20.83\rightarrow31.10$) while detection and
localization change less (\deta{} $39.57\rightarrow40.26$, \loca{}
$75.17\rightarrow75.18$). The small \deta{} change arises because the ID
runtime discards low-confidence newborns, whereas native IDs retain detections
at the detector threshold. Compared with the two leading entries, our largest
remaining deficit is likewise association. \assa{} is 31.10, while \loca{}
is within one point of second place.

\subsection{Ablation Studies}
\label{sec:exp_components}

Our ablations isolate three design questions. First, we test whether ID
prediction improves over the detector's native instance-bank identities and
whether its training should remain decoupled from the detector. The decoupled
variant freezes the detector while optimizing the ID modules. The
continued-detector variant updates both modules after ID convergence while
keeping their loss gradients decoupled.
Second, we vary the ID-token source and position encoding to test whether
performance depends on a particular feature composition. Third, we disable the
spatial gate and newborn recovery independently to distinguish the learned
decoder's contribution from explicit geometric constraints. The native versus
decoupled comparison and the latter two studies fix the upstream detector.
The retained output can still vary slightly because newborn score filtering
depends on the predicted identity. Table~\ref{tab:val_main} reports \deta{}
explicitly because continued detector training also alters the detections.

\begin{table}[t]
  \centering
  \caption{\textbf{Native identities and detector--ID training strategy.}
  Full 9{,}000-frame validation sequences at threshold 0.4. Native-ID rows use the
  corresponding detector's instance-bank identities. ``Decoupled'' freezes
  the submitted detector while training the ID modules. ``Continued det.''
  resumes the converged ID model and updates both modules with ID-loss
  gradients detached from the detector. Best per scene and metric in bold.}
  \label{tab:val_main}
  \setlength{\tabcolsep}{3.2pt}
  \resizebox{\textwidth}{!}{%
  \begin{tabular}{@{}l ccc ccc ccc@{}}
    \toprule
    & \multicolumn{3}{c}{\texttt{Warehouse\_020}}
    & \multicolumn{3}{c}{\texttt{Warehouse\_021}}
    & \multicolumn{3}{c}{\texttt{Warehouse\_022}} \\
    \cmidrule(lr){2-4}\cmidrule(lr){5-7}\cmidrule(lr){8-10}
    System & \hota{} & \deta{} & \assa{}
           & \hota{} & \deta{} & \assa{}
           & \hota{} & \deta{} & \assa{} \\
    \midrule
    Submitted frozen det.\ (native IDs)
      & 41.06 & 49.38 & 37.68
      &  5.98 & 10.33 &  4.20
      & 13.43 & \textbf{28.19} &  6.56 \\
    \quad $+$ ID pred.\ (decoupled)
      & \textbf{45.90} & \textbf{49.48} & \textbf{44.50}
      &  7.15 & 10.21 &  5.48
      & 14.38 & 28.05 &  7.54 \\
    \quad $+$ ID pred.\ (continued det.)
      & 31.05 & 33.43 & 30.71
      & 15.32 & 16.70 & 15.22
      & \textbf{14.52} & 26.71 & \textbf{8.50} \\
    \midrule
    Finetuned detector (native IDs)
      & 26.11 & 26.09 & 29.30
      & 16.16 & 16.45 & 21.81
      &  8.02 & 12.97 &  6.17 \\
    \quad $+$ ID pred.\ (decoupled)
      & 27.16 & 26.32 & 30.98
      & 16.20 & 16.73 & 18.71
      &  9.23 & 14.55 &  6.89 \\
    \quad $+$ ID pred.\ (continued det.)
      & 25.23 & 25.47 & 28.21
      & \textbf{19.54} & \textbf{16.82} & \textbf{30.12}
      & 10.20 & 17.05 &  6.44 \\
    \bottomrule
  \end{tabular}}
\end{table}

Table~\ref{tab:val_main} first compares native instance-bank identities with
decoupled ID prediction at a fixed detector. ID prediction improves \hota{} on
all three scenes for both detector initializations, $+4.84$, $+1.17$, and
$+0.95$ points on \texttt{Warehouse\_020}, \texttt{Warehouse\_021}, and
\texttt{Warehouse\_022} from the submitted frozen detector, and $+1.05$,
$+0.04$, and $+1.21$ points from the finetuned one. For the submitted detector,
the improvement is attributable to association, since freezing holds
\deta{} to within $0.14$ points while \assa{} rises by $6.82$, $1.28$, and
$0.98$ points. The finetuned rows do not decompose as cleanly, since \assa{}
falls from $21.81$ to $18.71$ on \texttt{Warehouse\_021}. The association
benefit is therefore consistent under the submitted frozen detector but not
universal.

The paired baseline uses the exact detector weights stored in the submitted
checkpoint. They originate from the public NGC model but include one warm-up
update before freezing (maximum parameter difference $9.5{\times}10^{-7}$).
A clean rerun reproduces both paired rows to 0.01 on every reported metric.
A separate raw-NGC rerun gives native-ID \hota{} of 38.20, 5.97, and 13.47.
We retain the exact-weight baseline so that its difference from the decoupled
row isolates the association layer.

The same table then compares decoupled and continued-detector training. The
latter gives its largest gains on \texttt{Warehouse\_021}, the scene the
submitted detector handles worst ($+8.17$ and $+3.34$ \hota{} for the submitted
and finetuned initializations), and smaller gains on \texttt{Warehouse\_022}
($+0.14$ and $+0.97$), but hurts \texttt{Warehouse\_020}, the scene it handles
best ($-14.85$ and $-1.93$). Three validation scenes, effectively two because
\texttt{Warehouse\_020} and \texttt{Warehouse\_021} differ only in rendering,
cannot settle which strategy generalizes. We therefore keep the detector
frozen in the submitted system.

\begin{table}[t]
  \centering
  \caption{\textbf{ID-token source and position encoding.} Frozen
  recurrent detector, geometric constraints enabled, threshold 0.3, and full
  9{,}000-frame validation sequences. The first three rows use matched 2k-iteration decoders. The
  deployed combined model was trained for 6k iterations and is shown for
  context, not as part of the controlled comparison. \deta{} varies by at most
  $0.6$ across rows on every scene and is omitted. Best among
  the matched 2k rows in bold.}
  \label{tab:tokens}
  \begin{tabular}{@{}llcccccc@{}}
    \toprule
    Token source & Encoding
      & \multicolumn{2}{c}{\texttt{W020}}
      & \multicolumn{2}{c}{\texttt{W021}}
      & \multicolumn{2}{c}{\texttt{W022}} \\
    \cmidrule(lr){3-4}\cmidrule(lr){5-6}\cmidrule(lr){7-8}
      & & \hota{} & \assa{} & \hota{} & \assa{} & \hota{} & \assa{} \\
    \midrule
    Detector query & none
      & \textbf{45.57} & \textbf{43.98}
      & \textbf{7.19} & \textbf{5.51}
      & \textbf{14.32} & \textbf{7.58} \\
    3D position & Fourier
      & 45.34 & 43.64 & 6.87 & 5.22 & 14.10 & 7.34 \\
    3D position & linear
      & 41.92 & 37.91 & 6.88 & 5.25 & 14.28 & 7.43 \\
    \midrule
    Query $+$ position & Fourier
      & 45.94 & 44.69 & 7.27 & 5.55 & 14.54 & 7.65 \\
    \bottomrule
  \end{tabular}
\end{table}

In the matched 2k comparison, the detector-query token is strongest on all three
scenes (Table~\ref{tab:tokens}). Its margin over Fourier-encoded position alone
is nevertheless small, $0.23$, $0.32$, and $0.22$~\hota{} on
\texttt{W020}, \texttt{W021}, and \texttt{W022}. Encoding sensitivity is not
consistent across scenes. Replacing the Fourier position code with a learnable
linear map costs $3.42$~\hota{} and $5.73$~\assa{} on \texttt{W020}, but
changes both metrics by at most $0.18$ on the other scenes. The deployed 6k
combined checkpoint has the highest observed values, but its additional
training prevents attributing those differences to token composition. We
therefore retain it as the submitted configuration and restrict this table
to within-table comparisons at threshold 0.3, rather than treating it as
evidence that query and position features are complementary.

\begin{table}[t]
  \centering
  \caption{\textbf{Geometric constraints} on the frozen-detector ID model
  (full 9{,}000-frame validation sequences, threshold 0.4). \deta{} varies by at most $0.09$ across
  configurations on every scene and is omitted.}
  \label{tab:gates}
  \begin{tabular}{@{}cccccccc@{}}
    \toprule
    Spatial & Newborn
      & \multicolumn{2}{c}{\texttt{W020}}
      & \multicolumn{2}{c}{\texttt{W021}}
      & \multicolumn{2}{c}{\texttt{W022}} \\
    \cmidrule(lr){3-4}\cmidrule(lr){5-6}\cmidrule(lr){7-8}
    gate & recovery
      & \hota{} & \assa{} & \hota{} & \assa{} & \hota{} & \assa{} \\
    \midrule
    \xmark & \xmark & 44.31 & 41.37 & 7.11 & 5.32 & 12.57 & 5.77 \\
    \cmark & \xmark & 45.85 & 44.29 & \textbf{7.35} & \textbf{5.68} & 11.46 & 4.82 \\
    \xmark & \cmark & 45.66 & 44.21 & 7.04 & 5.23 & 14.24 & 7.38 \\
    \cmark & \cmark & \textbf{45.90} & \textbf{44.50} & 7.15 & 5.48 & \textbf{14.38} & \textbf{7.54} \\
    \bottomrule
  \end{tabular}
\end{table}

Combining Tables~\ref{tab:val_main} and~\ref{tab:gates} isolates the complete
progression under the same frozen submitted detector and threshold 0.4. Native IDs,
unconstrained ID prediction, and constrained ID prediction respectively yield
$41.06\!\rightarrow\!44.31\!\rightarrow\!45.90$ \hota{} on \texttt{W020},
$5.98\!\rightarrow\!7.11\!\rightarrow\!7.15$ on \texttt{W021}, and
$13.43\!\rightarrow\!12.57\!\rightarrow\!14.38$ on \texttt{W022}. Thus the
learned decoder alone improves two scenes, while the geometric runtime is
necessary to improve over native IDs on \texttt{W022}.

The constraints act almost entirely on association, with \deta{} varying by at most
$0.09$ in Table~\ref{tab:gates}. Their usefulness, however, is scene-dependent.
On \texttt{W020} either constraint recovers most of the gain,
$+1.54$~\hota{} from the gate and $+1.35$ from recovery, versus $+1.59$
together. On the sparse \texttt{W022} they are not interchangeable. The gate alone
\emph{reduces} \hota{} by $1.11$, whereas newborn recovery alone recovers nearly
the full $+1.81$ gain. Localization is much weaker there
(\loca{}~46 \vs~77). The gate's degradation is consistent with localization
error causing a fixed metric threshold to reject genuine re-claims, while
newborn recovery can reconnect detections after assignment. On
\texttt{W021} all effects are within $0.25$~\hota{} of the unconstrained
model. The deployed combination is best or near-best on all three scenes,
and newborn recovery is more consistent than the fixed spatial gate.

\subsection{Limitations}
\label{sec:exp_analysis}

\begin{figure}[t]
  \centering
  \includegraphics[width=0.82\linewidth]{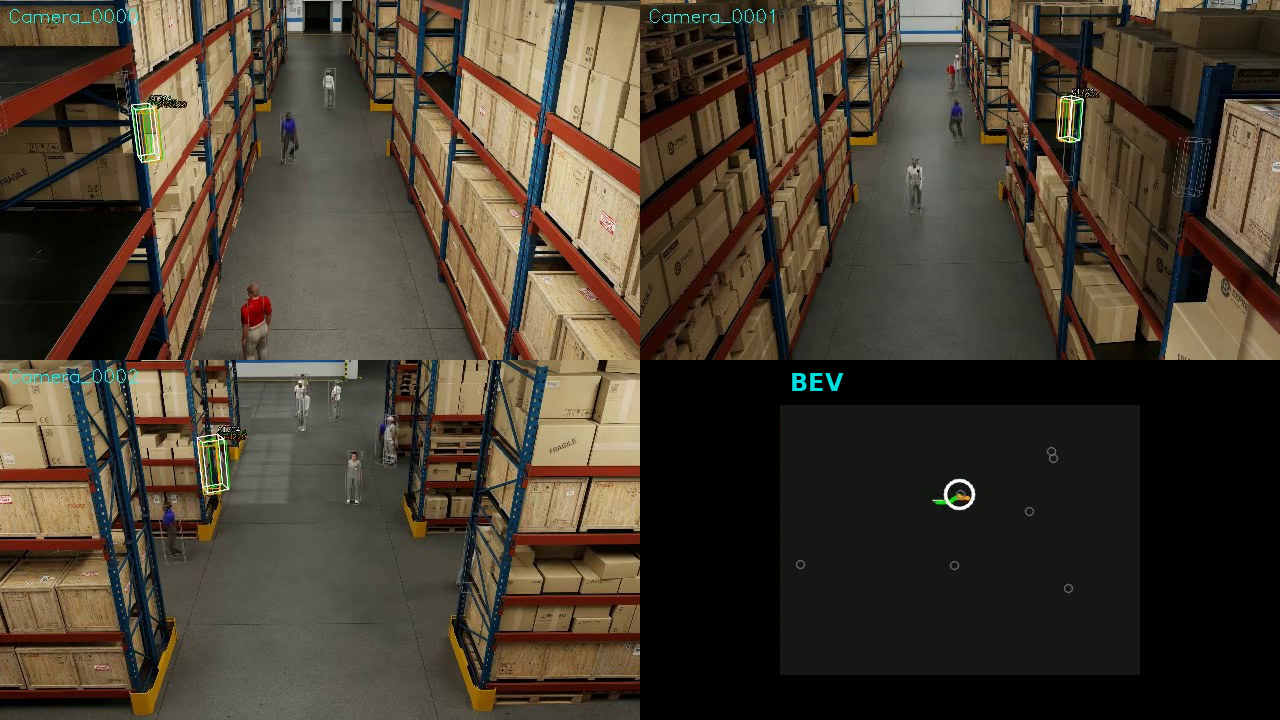}
  \caption{\textbf{Short-term identity confusion.} Three synchronized camera
  views and the bottom-right BEV show a close crossing on
  \texttt{Warehouse\_022}. One continuously observed person's predicted
  identity changes from 245 (green) to 229 (orange), while the nearest other
  track is 0.1\,m away. White denotes the corresponding ground-truth
  trajectory.}
  \label{fig:switch}
\end{figure}

The dominant residual error of the proposed method is associative rather than
positional (Sec.~\ref{sec:exp_main}). To characterize it, we match predictions
to ground truth frame by frame with a 1\,m ground-plane gate. Person errors are
observed as short-lived confusions when two trajectories approach within the
gate radius and separate again (Fig.~\ref{fig:switch}). Inspected vehicle and
robot errors also follow detection gaps exceeding the decoder's 29-frame
history. These qualitative examples expose two properties of the architecture.
The ID token is built from a detector query optimized for classification and
3D box regression rather than from a feature trained specifically for
re-identification, making spatially adjacent instances of the same class
harder to distinguish. Moreover, assignment is restricted to identities
observed within the history window, so a trajectory whose final token has left
that window necessarily receives a new identity. A dedicated appearance
embedding could address the first failure mode, but recovery beyond the window
would additionally require a longer or compressed identity memory. We leave
both extensions to future work.

The frozen detector still upper-bounds association quality. Across
\texttt{W020} and \texttt{W021}, which share geometry and identities but
differ in rendered appearance, pretrained \deta{} falls from 49.38 to 10.33.

\section{Conclusion}
\label{sec:conclusion}

We presented an online multi-camera 3D tracking architecture that
couples recurrent, globally fused sparse detections with causal ID
prediction. We retain MOTIP's relative-ID decoder and recycled-slot runtime,
adapt them to recurrent world-frame 3D observations, and add metric spatial
gating and proximity-based newborn recovery. On the official test set, adding
this association layer to the frozen detector raises \hota{} from 29.63 to
38.01. Validation shows that decoupled ID training improves \hota{} over native
instance-bank identities. The remaining errors motivate a dedicated appearance
embedding for close crossings and identity memory that outlasts the decoder's
temporal window.

\FloatBarrier
\bibliographystyle{splncs04}
\bibliography{main}

@inproceedings{motip,
  author    = {Gao, Ruopeng and Qi, Ji and Wang, Limin},
  title     = {Multiple Object Tracking as {ID} Prediction},
  booktitle = {CVPR},
  year      = {2025},
  pages     = {27883--27893}
}

@article{nvidia_omni3d,
  author  = {Wang, Yizhou and Pusegaonkar, Sameer and Wang, Yuxing and Li, Anqi and Kumar, Vishal and Sethi, Chetan and Aiyer, Ganapathy and He, Yun and Thakkar, Kartikay and Rathi, Swapnil and Rupde, Bhushan and Tang, Zheng and Biswas, Sujit},
  title   = {A Unified 3D Object Perception Framework for Real-Time Outside-In Multi-Camera Systems},
  journal = {arXiv preprint arXiv:2601.10819},
  year    = {2026}
}

@article{sparse4dv1,
  author  = {Lin, Xuewu and Lin, Tianwei and Pei, Zixiang and Huang, Lichao and Su, Zhizhong},
  title   = {Sparse4D: Multi-view 3D Object Detection with Sparse Spatial-Temporal Fusion},
  journal = {arXiv preprint arXiv:2211.10581},
  year    = {2022}
}

@article{sparse4dv2,
  author  = {Lin, Xuewu and Lin, Tianwei and Pei, Zixiang and Huang, Lichao and Su, Zhizhong},
  title   = {Sparse4D v2: Recurrent Temporal Fusion with Sparse Model},
  journal = {arXiv preprint arXiv:2305.14018},
  year    = {2023}
}

@article{sparse4dv3,
  author  = {Lin, Xuewu and Pei, Zixiang and Lin, Tianwei and Huang, Lichao and Su, Zhizhong},
  title   = {Sparse4D v3: Advancing End-to-End 3D Detection and Tracking},
  journal = {arXiv preprint arXiv:2311.11722},
  year    = {2023}
}

@inproceedings{detr,
  author    = {Carion, Nicolas and Massa, Francisco and Synnaeve, Gabriel and Usunier, Nicolas and Kirillov, Alexander and Zagoruyko, Sergey},
  title     = {End-to-End Object Detection with Transformers},
  booktitle = {ECCV},
  year      = {2020}
}

@inproceedings{deformabledetr,
  author    = {Zhu, Xizhou and Su, Weijie and Lu, Lewei and Li, Bin and Wang, Xiaogang and Dai, Jifeng},
  title     = {Deformable {DETR}: Deformable Transformers for End-to-End Object Detection},
  booktitle = {ICLR},
  year      = {2021}
}

@inproceedings{motr,
  author    = {Zeng, Fangao and Dong, Bin and Zhang, Yuang and Wang, Tiancai and Zhang, Xiangyu and Wei, Yichen},
  title     = {{MOTR}: End-to-End Multiple-Object Tracking with Transformer},
  booktitle = {ECCV},
  year      = {2022}
}

@inproceedings{mutr3d,
  author    = {Zhang, Tianyuan and Chen, Xuanyao and Wang, Yue and Wang, Yilun and Zhao, Hang},
  title     = {{MUTR3D}: A Multi-Camera Tracking Framework via 3D-to-2D Queries},
  booktitle = {CVPR Workshops},
  year      = {2022},
  pages     = {4537--4546}
}

@inproceedings{pftrack,
  author    = {Pang, Ziqi and Li, Jie and Tokmakov, Pavel and Chen, Dian and Zagoruyko, Sergey and Wang, Yu-Xiong},
  title     = {Standing Between Past and Future: Spatio-Temporal Modeling for Multi-Camera 3D Multi-Object Tracking},
  booktitle = {CVPR},
  year      = {2023},
  pages     = {17928--17938}
}

@inproceedings{dqtrack,
  author    = {Li, Yanwei and Yu, Zhiding and Philion, Jonah and Anandkumar, Anima and Fidler, Sanja and Jia, Jiaya and Alvarez, Jose},
  title     = {End-to-End 3D Tracking with Decoupled Queries},
  booktitle = {ICCV},
  year      = {2023},
  pages     = {18302--18311}
}

@inproceedings{adatrack,
  author    = {Ding, Shuxiao and Schneider, Lukas and Cordts, Marius and Gall, Juergen},
  title     = {{ADA-Track}: End-to-End Multi-Camera 3D Multi-Object Tracking with Alternating Detection and Association},
  booktitle = {CVPR},
  year      = {2024},
  pages     = {15184--15194}
}

@inproceedings{trackformer,
  author    = {Meinhardt, Tim and Kirillov, Alexander and Leal-Taix{\'e}, Laura and Feichtenhofer, Christoph},
  title     = {TrackFormer: Multi-Object Tracking with Transformers},
  booktitle = {CVPR},
  year      = {2022}
}

@inproceedings{memot,
  author    = {Cai, Jiarui and Xu, Mingze and Li, Wei and Xiong, Yuanjun and Xia, Wei and Tu, Zhuowen and Soatto, Stefano},
  title     = {{MeMOT}: Multi-Object Tracking with Memory},
  booktitle = {CVPR},
  year      = {2022}
}

@inproceedings{bytetrack,
  author    = {Zhang, Yifu and Sun, Peize and Jiang, Yi and Yu, Dongdong and Weng, Fucheng and Yuan, Zehuan and Luo, Ping and Liu, Wenyu and Wang, Xinggang},
  title     = {ByteTrack: Multi-Object Tracking by Associating Every Detection Box},
  booktitle = {ECCV},
  year      = {2022}
}

@inproceedings{deepsort,
  author    = {Wojke, Nicolai and Bewley, Alex and Paulus, Dietrich},
  title     = {Simple Online and Realtime Tracking with a Deep Association Metric},
  booktitle = {ICIP},
  year      = {2017},
  pages     = {3645--3649}
}

@inproceedings{ocsort,
  author    = {Cao, Jinkun and Pang, Jiangmiao and Weng, Xinshuo and Khirodkar, Rawal and Kitani, Kris},
  title     = {Observation-Centric {SORT}: Rethinking {SORT} for Robust Multi-Object Tracking},
  booktitle = {CVPR},
  year      = {2023}
}

@inproceedings{tracktrack,
  author    = {Shim, Kyujin and Ko, Kangwook and Yang, Yujin and Kim, Changick},
  title     = {Focusing on Tracks for Online Multi-Object Tracking},
  booktitle = {CVPR},
  year      = {2025},
  pages     = {11687--11696}
}

@article{hota,
  author  = {Luiten, Jonathon and O{\v{s}}ep, Aljo{\v{s}}a and Dendorfer, Patrick and Torr, Philip and Geiger, Andreas and Leal-Taix{\'e}, Laura and Leibe, Bastian},
  title   = {{HOTA}: A Higher Order Metric for Evaluating Multi-Object Tracking},
  journal = {IJCV},
  volume  = {129},
  pages   = {548--578},
  year    = {2021}
}

@inproceedings{wildtrack,
  author    = {Chavdarova, Tatjana and Baqu{\'e}, Pierre and Bouquet, St{\'e}phane and Maksai, Andrii and Jose, Cijo and Bagautdinov, Timur and Lettry, Louis and Fua, Pascal and Van Gool, Luc and Fleuret, Fran{\c{c}}ois},
  title     = {{WILDTRACK}: A Multi-camera {HD} Dataset for Dense Unscripted Pedestrian Detection},
  booktitle = {CVPR},
  year      = {2018}
}

@inproceedings{mvdet,
  author    = {Hou, Yunzhong and Zheng, Liang and Gould, Stephen},
  title     = {Multiview Detection with Feature Perspective Transformation},
  booktitle = {ECCV},
  year      = {2020}
}

@inproceedings{mvdetr,
  author    = {Hou, Yunzhong and Zheng, Liang},
  title     = {Multiview Detection with Shadow Transformer (and View-Coherent Data Augmentation)},
  booktitle = {ACM Multimedia},
  year      = {2021}
}

@inproceedings{earlybird,
  author    = {Teepe, Torben and Wolters, Philipp and Gilg, Johannes and Herzog, Fabian and Rigoll, Gerhard},
  title     = {EarlyBird: Early-Fusion for Multi-View Tracking in the Bird's Eye View},
  booktitle = {WACV Workshops},
  year      = {2024}
}

@inproceedings{lmgp,
  author    = {Nguyen, Duy M. H. and Henschel, Roberto and Rosenhahn, Bodo and Sonntag, Daniel and Swoboda, Paul},
  title     = {{LMGP}: Lifted Multicut Meets Geometry Projections for Multi-Camera Multi-Object Tracking},
  booktitle = {CVPR},
  year      = {2022},
  pages     = {8866--8875}
}

@inproceedings{tracktacular,
  author    = {Teepe, Torben and Wolters, Philipp and Gilg, Johannes and Herzog, Fabian and Rigoll, Gerhard},
  title     = {Lifting Multi-View Detection and Tracking to the Bird's Eye View},
  booktitle = {CVPR Workshops},
  year      = {2024},
  pages     = {667--676}
}

@inproceedings{rest,
  author    = {Cheng, Cheng-Che and Qiu, Min-Xuan and Chiang, Chen-Kuo and Lai, Shang-Hong},
  title     = {{ReST}: A Reconfigurable Spatial-Temporal Graph Model for Multi-Camera Multi-Object Tracking},
  booktitle = {ICCV},
  year      = {2023},
  pages     = {10051--10060}
}

@inproceedings{depthtrack,
  author    = {Tran, Tai Huu-Phuong and Tran, Duong Nguyen-Ngoc and Huynh, Ngoc Doan-Minh and Tran, Chi Dai and Pham, Long Hoang and Ho, Quoc Pham-Nam and Nguyen, Huy-Hung and Vu, Duong Khac and Jeon, Hyung-Min and Jeon, Hyung-Joon and Phan, Son Hong and Le Ba Khanh, Trinh and Jeon, Jae Wook},
  title     = {DepthTrack: Cluster Meets {BEV} for Multi-Camera Multi-Target 3D Tracking},
  booktitle = {ICCV Workshops},
  year      = {2025},
  pages     = {5348--5357}
}

@inproceedings{mcblt,
  author    = {Wang, Yizhou and Meinhardt, Tim and Cetintas, Orcun and Yang, Cheng-Yen and Pusegaonkar, Sameer and Missaoui, Benjamin and Biswas, Sujit and Tang, Zheng and Leal-Taix{\'e}, Laura},
  title     = {{MCBLT}: Multi-Camera Multi-Object 3D Tracking in Long Videos},
  booktitle = {ICCV Workshops},
  year      = {2025},
  pages     = {5304--5313}
}

@inproceedings{gmt,
  author    = {Zhen, Yihao and Xu, Mingyue and Wang, Qiang and Fan, Baojie and Dong, Jiahua and Zhao, Tinghui and Fan, Huijie},
  title     = {{GMT}: Effective Global Framework for Multi-Camera Multi-Target Tracking},
  booktitle = {CVPR},
  year      = {2026},
  pages     = {28201--28210}
}

@inproceedings{Tang26AICity26,
  author    = {Tang, Zheng and Wang, Shuo and Anastasiu, David C. and Chang, Ming-Ching and others},
  title     = {The 10th {AI} City Challenge},
  booktitle = {ECCV Workshops},
  year      = {2026},
  address   = {Malm{\"o}, Sweden}
}

@inproceedings{aicity2025,
  author    = {Tang, Zheng and Wang, Shuo and Anastasiu, David C. and Chang, Ming-Ching and Sharma, Anuj and Kong, Quan and Kobori, Norimasa and Gochoo, Munkhjargal and Batnasan, Ganzorig and Otgonbold, Munkh-Erdene and Alnajjar, Fady and Hsieh, Jun-Wei and Kornuta, Tomasz and Li, Xiaolong and Zhao, Yilin and Zhang, Han and Radhakrishnan, Subhashree and Jain, Arihant and Kumar, Ratnesh and Murali, Vidya N. and Wang, Yuxing and Pusegaonkar, Sameer Satish and Wang, Yizhou and Biswas, Sujit and Wu, Xunlei and Zheng, Zhedong and Chakraborty, Pranamesh and Chellappa, Rama},
  title     = {The 9th {AI} City Challenge},
  booktitle = {ICCV Workshops},
  year      = {2025},
  pages     = {5526--5535}
}

\end{document}